\documentclass[pmlr]{jmlr}
\usepackage[utf8]{inputenc}

\usepackage{fnpct}
\usepackage{graphicx}
\usepackage{booktabs}
\usepackage{multirow}
\usepackage{amsmath, amsfonts, bm}
\usepackage{soul} 
\usepackage{lipsum}
\usepackage{mathtools}
\usepackage[shortlabels]{enumitem}

\newcommand{\qerr}{q_\text{err}}
\newcommand{\defeq}{\vcentcolon=}

\jmlrvolume{204}
\jmlryear{2023}
\jmlrworkshop{Conformal and Probabilistic Prediction with Applications}
\jmlrproceedings{PMLR}{Proceedings of Machine Learning Research}

\title{Evaluating Machine Translation Quality with Conformal Predictive Distributions}

\author{
    \Name{Patrizio Giovannotti} \Email{patrizio.giovannotti.2019@live.rhul.ac.uk}\\
    \addr{Royal Holloway, University of London, Egham, Surrey, UK}\\
    \addr{Centrica plc, UK}
}

\editor{}

\begin{document}

\maketitle

\begin{abstract}
This paper presents a new approach for assessing uncertainty in machine translation by simultaneously evaluating translation quality and providing a reliable confidence score. Our approach utilizes conformal predictive distributions to produce prediction intervals with guaranteed coverage, meaning that for any given significance level $\epsilon$, we can expect the true quality score of a translation to fall out of the interval at a rate of $1 - \epsilon$. In this paper, we demonstrate how our method outperforms a simple, but effective baseline on six different language pairs in terms of coverage and sharpness. Furthermore, we validate that our approach requires the data exchangeability assumption to hold for optimal performance.
\end{abstract}


\section{Introduction}
Machine translation (MT) is the task of automatically translating text from one language to another using a computer program. With the growing globalization of businesses and the increasing availability of multilingual content on the internet, machine translation has become an essential technology for communication across language barriers. However, machine translation is still far from perfect and often produces translations that are inaccurate or of poor quality. Evaluating the quality of machine-translated text is therefore crucial for ensuring the usability and effectiveness of translation systems. 
Among the several ways to evaluate MT outputs, quality estimation (QE) is the task of determining a quality score for machine-translated text without the help of reference translations. Good quality estimation is essential whenever a real-time decision must be made about a translation output, like whether to publish a translated text or informing the user about the confidence in a translation. 
QE can also be useful for other purposes such as selecting the best translation among multiple candidates or providing feedback to MT developers.

Since QE was introduced as a shared task in the Conference of Machine Translation (\citealp{callison-burch-etal-2012-findings}), increasingly more systems have been able to improve on the baselines and redefine the state of the art. However, few efforts have been made towards refining how these models quantify the uncertainty of their predictions. In this paper, we propose a novel approach to express the uncertainty of quality estimates with the help of conformal predictive distributions (CPD -- \citealp{pmlr-v60-vovk17a}). Under the sole assumption of the data being IID, our method produces prediction intervals with \textit{guaranteed coverage}, that is, for any chosen confidence level $1-\epsilon$, prediction intervals will fail to include the correct label at a rate $\epsilon$. The ability to specify arbitrary intervals can be useful in many ways. For example, we can set a quality threshold $q_\text{err}$ and compute the probability of a proposed translation to have quality smaller than $\qerr$, i.e. to be ``not good enough''. CPD ensures that such a probability reflects the long-term relative frequency of correct predictions. For example, let the random variable $Q$ model the quality score of any sentence pair; let $\mathcal{S}$ be the set of examples where our model predicts $P(Q\leq \qerr)=0.8$. Then, the number of examples in $\mathcal{S}$ for which the true label is smaller than $\qerr$ is $|\mathcal{S}|\cdot0.8$.

Our contributions are the following: we demonstrate how to apply CPD to state-of-the-art MT evaluation models to equip them with uncertainty estimation capabilities; we verify that we can build prediction intervals with good coverage, given that the IID assumption holds, for several language pairs; we report results on the ability of these models to predict translation failures, that is where their quality score is too low to be accepted.


\section{Related Work}
In this section, we provide some context on the quality estimation problem and prior studies that have focused on estimating its uncertainty. Moreover, we discuss the application of conformal methods to various areas of natural language processing (NLP).
\subsection{Background}
Following \citet{glushkova-etal-2021-uncertainty-aware}, we can formalise our problem as follows: let $s$ be a sentence in the source language, $t$ the same sentence translated by an MT system, and $\mathcal{R}=\{r_1, \dots, r_{|\mathcal{R}|}\}$ a set of reference translations. An MT evaluator is a function that accepts as input a tuple $\langle s, t, \mathcal{R}\rangle$ and outputs a quality score $\hat q\in\mathbb{R}$.

The simplest evaluator would be a metric that could quantify the amount of lexical overlap between $t$ and $r_i$. Examples of this approach are the popular BLEU score (\citealp{bleu-papineni-2002}) and \textsc{Meteor} (\citealp{lavie2009meteor}). The success of neural machine translation techniques inspired new metrics such as \textsc{BERTScore} \citep{Zhang-2020-BERTScore:}, where quality scores depends on the pairwise cosine similarity between words in $t$ and $r_i$, once each word is encoded via a BERT model (\citealp{devlin-etal-2019-bert}). Despite BERT-based metrics showed to be robust under distribution shifts \citep{vu-etal-2022-layer} it is still unclear how much these metrics are suitable to adequately reflect a model's performance \citep{blagec-etal-2022-global}. 

The evaluation metrics mentioned above require a set of reference translations. The field of \textit{quality estimation} is focused on the case $\mathcal{R} = \O$, i.e. estimating translation quality in the absence of reference translations.


\subsection{Quality estimation for MT}
In many real-world situations, reference translations are not available, and the only score returned by modern transformer-based models is the sum of the log-probabilities of each token in the generated sentence. This score, denoted by a real number $c\in(-\infty, 0]$, can be used to rank different translation candidates, however it does not generally correlate with human judgement concerning quality of translation.

To help in this scenario, known as quality estimation, several datasets with human-annotated quality scores have been created. Such scores include direct assessments (DA, \citealp{graham-etal-2013-continuous-DA}) and human translation error rates (HTER, \citealp{snover-etal-2006-study-hter}). MT evaluation systems are asked to generate quality scores $\hat q$ that correlate with ground truth scores $q^*$ as much as possible, in what is in essence a regression task. 

\textsc{Bleurt} \citep{sellam-etal-2020-bleurt} and \textsc{Comet} \citep{rei-etal-2020-comet} are two examples of quality estimators, with the latter relying on a multilingual RoBERTa pre-trained model \citep{liu2019roberta}.

\subsection{Uncertainty quantification in MT}
Although many QE efforts have achieved good predictive performance, a significant limitation of the proposed models is that they are often unable to convey the uncertainty associated with their predictions. The entire topic of uncertainty quantification for MT has not been explored enough, with \citet{beck-etal-2016-exploring} being among the very few to have addressed the issue in several years.

More recently, \citet{glushkova-etal-2021-uncertainty-aware} presented a modified version of \textsc{Comet} that outputs quality scores \textit{intervals} of variable width, depending on the confidence associated to the prediction. They propose to treat translation quality as a random variable $Q$ and predict a distribution $\hat P_Q(q)$, rather than a point estimate $\hat q$. They choose two parametric approaches: \textit{MC dropout}, where $h$ is run $N$ times, with different units dropped out each time, and \textit{deep ensembles}, where $N$ separate $h$ instances are randomly initialised and used to obtain scores. Both methods provide a set of quality scores $\mathcal{Q}=\{\hat q_1,\dots,\hat q_N\}$. The authors treat $\mathcal{Q}$ as a sample drawn from a Gaussian distribution, hence they estimate mean $\hat\mu$ and variance $\hat\sigma^2$ to use in the calculation of confidence intervals $I[q_{min}(\epsilon), q_{max}(\epsilon)]$ for $\epsilon\in[0,1]$. The aim is to provide a quality score interval that includes $q^*$ as much as possible, while being as tight as possible. Some limitations of this approach are the need of training several models or predict several times, the need for a post-calibration step and, more importantly, the strong assumption about the shape of $\hat P_Q(q)$. In contrast, our approach requires only one model, is well calibrated out of the box and makes no assumption about the distribution of $\hat P_Q(q)$.
 



\subsection{Conformal methods for NLP}
Conformal prediction has been applied in several forms to numerous NLP tasks. \citet{pmlr-v105-paisios19a} first explored the use of traditional CP for text classification, while \citet{pmlr-v128-maltoudoglou20a} and \citet{pmlr-v152-giovannotti21a} extended it to sentiment analysis and paraphrase detection by building on a BERT pre-trained model and experimenting with new nonconformity measures; \citet{pmlr-v179-giovannotti22a} studied the use of Venn--ABERS predictors \citep{Vovk2014VennAbersP} in the context of binary classification for natural language understanding; \citet{dey-2022-infilling} applied CP to part-of-speech tagging and the important task of text infilling (or masked language modelling). Very recently, \citet{robinson2023leveraging} examined the use of conformal prediction for question answering in the context of large language models, while \citet{ravfogel2023conformal} used a conformal approach to calibrate the parameter $p$ in top-$p$ sampling for language generation.

Conformal methods have also been used to optimize transformers, the current state-of-the-art of NLP architectures: \citet{schuster-calm-neurips-2022} proposed a method to accelerate text generation through an early-exiting mechanism enabled by conformal prediction.

\section{Methodology}
Our methods are rooted in recent advances in conformal prediction (CP), a machine learning framework introduced by \citet{gammerman1998} and fully developed by \citet{vovk2005algorithmic}.

\subsection{Conformal predictive distributions}
Introduced in \citet{pmlr-v60-vovk17a}, conformal predictive distributions (CPDs) are a novel approach to estimating the probability distribution of a continuous variable that depends on a number of features. Under minimal assumptions, i.e. the data being generated independently by an unknown fixed distribution, CPDs provide probabilities that correspond to long-term frequencies. CPDs make no assumption on the particular distribution of the data, hence no prior is required either.

In this work, we will use a computationally efficient version of CPD, namely split conformal predictive distributions \citep{VOVK2020292}. In the split CPDs framework, we require our original training sequence $z_1,\dots,z_n$ to be divided into a proper training sequence $z_1,\dots,z_m$ and a calibration sequence $z_{m+1},\dots,z_{n}$. Here each observation is a pair $z=(x,y)$ of an object $x\in\mathbf{X}$ and its label $y\in\mathbb{R}$, where $\mathbf{X}$ is any nonempty measurable space.

The essential component of CPDs is a (split) conformity measure, a function that should indicate how large is any label $y_{m+1}$ compared to the $m$ labels in the proper training set. The standard choice of conformity measure for a (test) observation $(x,y)$ is
\[
A(z_1,\dots,z_m,(x,y)) = \frac{y-\hat y}{\hat\sigma}
\]
where $\hat y$ is a prediction for $y$ and $\hat\sigma$ is an estimate of the quality of $\hat y$ (this is also referred as \textit{difficulty}, see also \citealp{pmlr-v152-bostrom21a-mondrian-cpd}). We will obtain $\hat y$ from a pretrained transformer model that we fine-tune on $z_1,\dots,z_m$ (see Section \ref{roberta}); as for $\hat\sigma$, we use the sum of the distances between $(x,y)$ and its $K$ nearest neighbours.

Once $A$ is defined, we are able to compute the conformity scores 
\begin{align*}
    \alpha_i &= A(z_1,\dots,z_m,(x_i,y_i)) \qquad i=m+1,\dots,n \\
    \alpha^y &= A(z_1,\dots,z_m,(x,y))
\end{align*}
which express how well a label conforms to the property of being large (we think of a label as nonconforming only when it is too small). The \textit{split conformal transducer} built on $A$ produces the values
\[
Q(z_1, \dots, z_{n}, (x,y))\defeq\frac{1}{n-m+1}|\{i=m+1,\dots,n \mid \alpha_i < \alpha^y\}| + Q_\tau
\]
with 
$$Q_\tau\defeq\frac{\tau}{n-m+1}|\{i=m+1,\dots,n\mid\alpha_i=\alpha^y\}| + \frac{\tau}{n-m+1}, \qquad \tau\sim U(0,1).$$

Because of the validity property of conformal transducers, values of $Q$ are uniformly distributed on $[0,1]$ when $z_1,\dots,z_m,z$ are IID.
Additionally, it can be proved that the function $Q$ is monotonically increasing in $y$ and tends asymptotically to 0 and 1 when $y$ tends to $-\infty$ and $+\infty$, respectively. These properties are what defines \textit{randomized predictive systems}, which are able to generate distribution functions. In other words, CPDs can be considered as p-values arranged into a distribution function (see Figure \ref{fig:example}).

An accessible tutorial on CPDs was written by \citet{toccaceli-conformal-predictive-distributions}.

\begin{figure}[htbp]
\floatconts
{fig:example}
{\caption{Conformal predictive distribution for a test example of the English→German dataset. Values for the quality label $y$ are normalized.}}
{\includegraphics[scale=0.55]{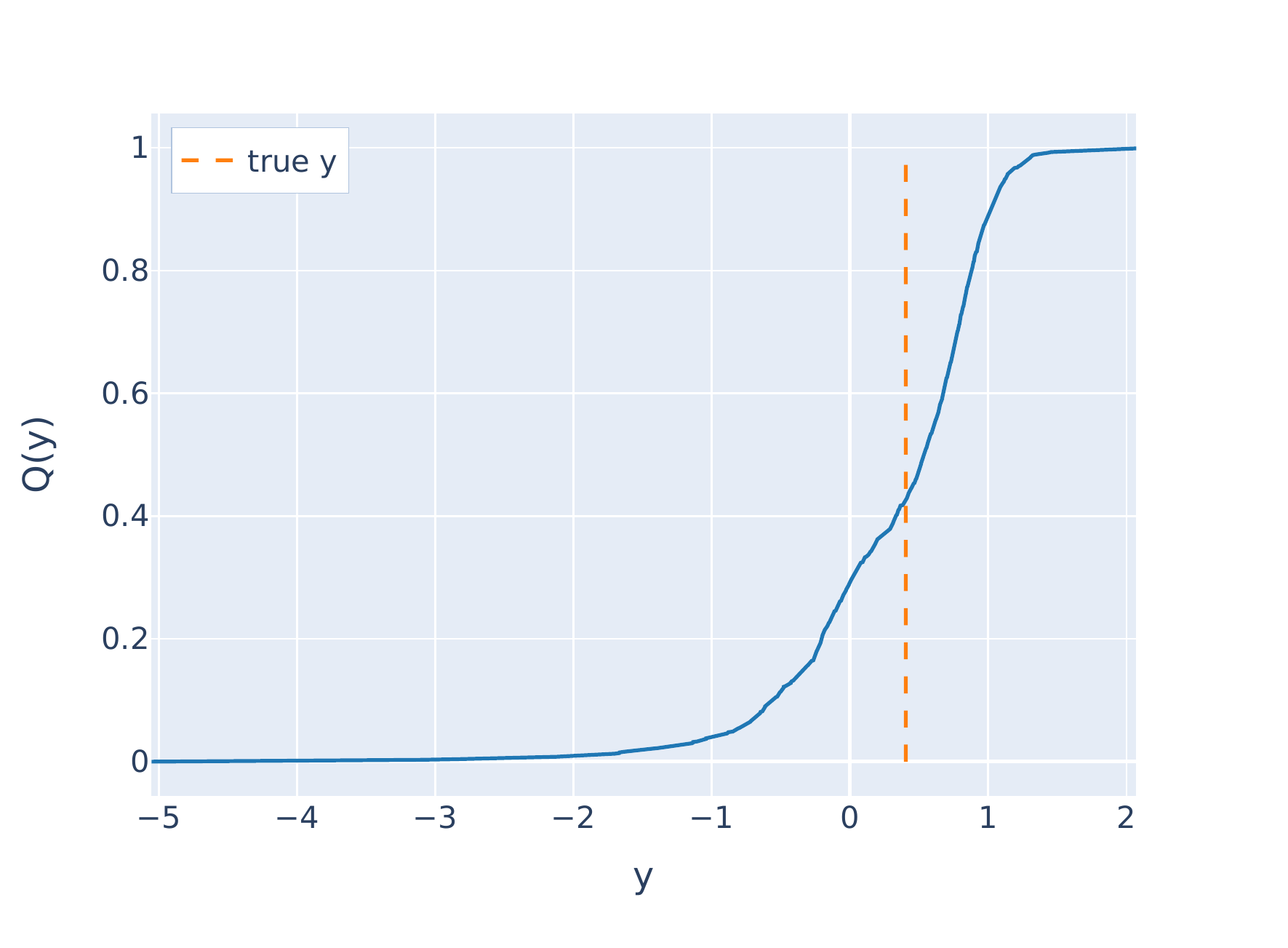}}
\end{figure}

In practice, for a new test object $x$, a split conformal predictive system can be implemented as follows: 
\begin{enumerate}
    \item Predict a label $\hat y\in\mathbb{R}$ and its estimated quality $\hat\sigma$ 
    \item Calculate the calibration scores $C_i=\hat y + \frac{\hat\sigma}{\hat\sigma_i}(y_i-\hat y_i)$ for $i={m+1},\dots,n$
    \item Sort the confidence scores in ascending order, obtaining the sequence $C_{(1)},\dots, C_{(n-m)}$, then set $C_{(0)}=-\infty$ and $C_{(n-m+1)}=\infty$
    \item Return the predictive distribution  
    \begin{equation}
        Q(y)=
        \begin{cases}
            \frac{i+\tau}{n-m+1} & \text{if } y\in(C_i, C_{i+1}) \text{, for } i\in\{0,1,\dots,n-m\}\\[10pt]
            \frac{i'-1+(i''-i'+2)\tau}{n-m+1} & \text{if } y=C_i \text{, for } i\in\{1,\dots,n-m\}
        \end{cases}
    \end{equation}
    where $i'\defeq\min\{j:C_j=C_i\}$ and $i''\defeq\max\{j:C_j=C_i\}$.
\end{enumerate}


\subsection{Underlying algorithm}\label{roberta}
Our algorithm of choice for predicting quality scores is XLM-RoBERTa \citep{conneau-etal-2020-unsupervised-xlm-roberta}, a multilingual masked language model that was specifically trained to compute dense representations of pairs of sentences written in different languages.

As a variant of the RoBERTa model \citep{liu2019roberta} pre-trained on multilingual corpora, XLM-RoBERTa is a transformer-based architecture that learns representations of text by processing it in a series of layers, each of which applies self-attention to the input tokens. This allows the model to capture long-range dependencies and contextual information in the text. The RoBERTa model was pre-trained on a large corpus of diverse text in multiple languages using the masked language modelling (MLM) and the next sentence prediction (NSP) objectives. This pre-training has been shown to improve the model's performance on downstream NLP tasks.

XLM-RoBERTa further improves upon the RoBERTa model by using cross-lingual language modeling (XLM) pre-training. This involves training the model on multiple languages simultaneously and encouraging it to learn shared representations of language. This enables the model to transfer knowledge between languages, which can be useful for MT quality estimation, as the model can learn to recognize patterns in one language that are indicative of high quality translations in another language.

In our experiments, we fine-tune the pre-trained \texttt{xlm-roberta-base} model -- equipped with a single-unit classification head, for regression tasks -- on the MT quality estimation dataset and use its predicted quality scores as a feature to train a second, lighter KNN model. Our KNN model takes as input these RoBERTa predictions and the quality scores predicted by another baseline model, which are included with in WMT 20 dataset. Training the KNN model over these two features resulted in improved Pearson's correlation with the true labels.





\subsection{Baseline}
We compare our approach with the same baseline method proposed by \citet{glushkova-etal-2021-uncertainty-aware}, which proved to perform reasonably well despite its simplicity. As a baseline, we map the original quality scores $\hat q$ computed by our RoBERTa model to a Gaussian distribution $\mathcal{N}(q;\hat\mu,\hat\sigma^2)$ with $\hat\mu\defeq\hat q$ and $\hat\sigma^2\defeq\sigma_\text{fixed}$, the average of the squared residuals $(\hat q - \hat\mu)^2$ over the validation set. One major limitation of our baseline is its inability to adapt the quality intervals' width to the uncertainty associated to each example. 

In order to generate the prediction interval for a confidence level $\alpha$, we compute $$[q_{\text{min}}(\alpha), q_{\text{max}}(\alpha)] = \hat\mu\pm\hat\sigma\cdot\text{probit}\left(\frac{1+\alpha}{2}\right)$$
 where we used the quantile function: $\text{probit}(p)=\sqrt{2}\text{erf}^{-1}(2p-1)$, where $\text{erf}$ is the error function.

\subsection{Evaluation Metrics} \label{eva}
We evaluated the performance of our proposed method, CPD, against our baseline approach across three metrics: Expected Calibration Error (ECE), Sharpness, and AUROC.


\paragraph{Expected Calibration Error (ECE)} ECE measures the difference between the expected accuracy of a set of predictions and their actual accuracy. It was first introduced by \citet{naeini2015obtaining}. We will use the version described by \citet{pmlr-v80-kuleshov18a} for regression:

$$\text{ECE} = \frac{1}{|\mathcal{E}|}\sum_{\epsilon\in\mathcal{E}} |\text{err}(\epsilon) - \epsilon|,$$

where $\mathcal{E}$ is a set of significance levels $\epsilon\in [0,1]$ and $\text{err}(\epsilon)$ is the error rate at a given significance level, namely the proportion of prediction intervals that do not include the true label (for a perfectly calibrated system, $\text{ECE}=0$). This type of error depends on the number of significance levels we consider: in our work, we chose $|\mathcal{E}|=50$, that is $\mathcal{E}=\{0.00, 0.02, 0.04, \dots, 1.00\}$.

\paragraph{Sharpness} Sharpness measures the degree of concentration of the predicted scores around the actual scores. A sharper distribution implies that the model is more confident in its predictions. In our work, sharpness will be measured by the average prediction interval width at a certain confidence level. We report sharpness values for prediction intervals at 90\% confidence, since we are more interested in high-confidence predictions, whereas higher-confidence intervals are likely to be too wide to be useful in a real scenario.
\paragraph{AUROC}
The Area Under the Receiver Operating Characteristic Curve measures the ability of a model to distinguish between positive and negative samples, and is particularly useful when the classes are imbalanced. AUROC measures the area under the curve obtained by plotting a model's true positive rate against its false positive rate at different classification thresholds. The AUROC score ranges from 0.5 to 1, with a score of 0.5 indicating random guessing and a score of 1 indicating perfect classification. We use AUROC to assess the ability of our models to detect critically wrong translations, i.e. with quality $q^*$ in the bottom decile of the test set.



\section{Experiments}
Our experiments made extensive use of the two main Hugging Face libraries \texttt{transformers} \citep{wolf-etal-2020-transformers} and \texttt{datasets} \citep{lhoest-etal-2021-datasets}. The conformal predictive distribution implementation used is \texttt{crepes} \citep{pmlr-v179-bostrom22a-crepes}.
\subsection{Datasets and experimental setup}
Our dataset was released for Task 1 of the WMT 2020 conference \citep{specia-etal-2020-findings-wmt}. It consists of labelled sentence pairs $(s, t)$ for 6 language pairs: two high-resource English→German (En-De) and English→Chinese (En-Zh) pairs; two medium-resource Romanian→English (Ro-En) and Estonian→English (Et-En) pairs; and two low-resource Sinhala→English (Si-En) and Nepali→English (Ne-En) pairs. Most of the sentences are extracted from Wikipedia and translated with a transformer-based NMT model trained using publicly available data. 

Each sentence is manually labelled with a score from 0 to 100 by a group of 3 independent annotators. The score is assigned on the basis of the following guidelines: the 0-10 range represents an incorrect translation; 11-29, a translation with few correct keywords, but the overall meaning is different from the source; 30-50, a translation with major mistakes; 51-69, a translation which is understandable and conveys the overall meaning of the source but contains typos or grammatical errors; 70-90, a translation that closely preserves the semantics of the source sentence; and 91-100, a perfect translation. These guidelines were introduced in FLORES \citep{guzman-etal-2019-flores}.

All quality scores are then transformed into their z-normalized values. The label to predict is then $y=\frac{q^*-\mu_\mathcal{D}}{\sigma_\mathcal{D}}$ where $\mu_\mathcal{D}, \sigma_\mathcal{D}$ are respectively the mean and the standard deviation of the quality scores $q^*$ in the dataset $\mathcal{D}$.
For more information about the dataset creation and composition refer to \citet{specia-etal-2020-findings-wmt}.

\begin{figure}[htbp]
\floatconts
    {fig:distro-original}
    {\caption{Distribution of label values (\emph{a}) before and (\emph{b}) after shuffling the Estonian→English dataset splits. Before shuffling, there is an evident mismatch between train and test distributions, hence the IID property may not hold.}}
    {%
        \subfigure{%
            \label{fig:distrib-original}
            \includegraphics[scale=.66]{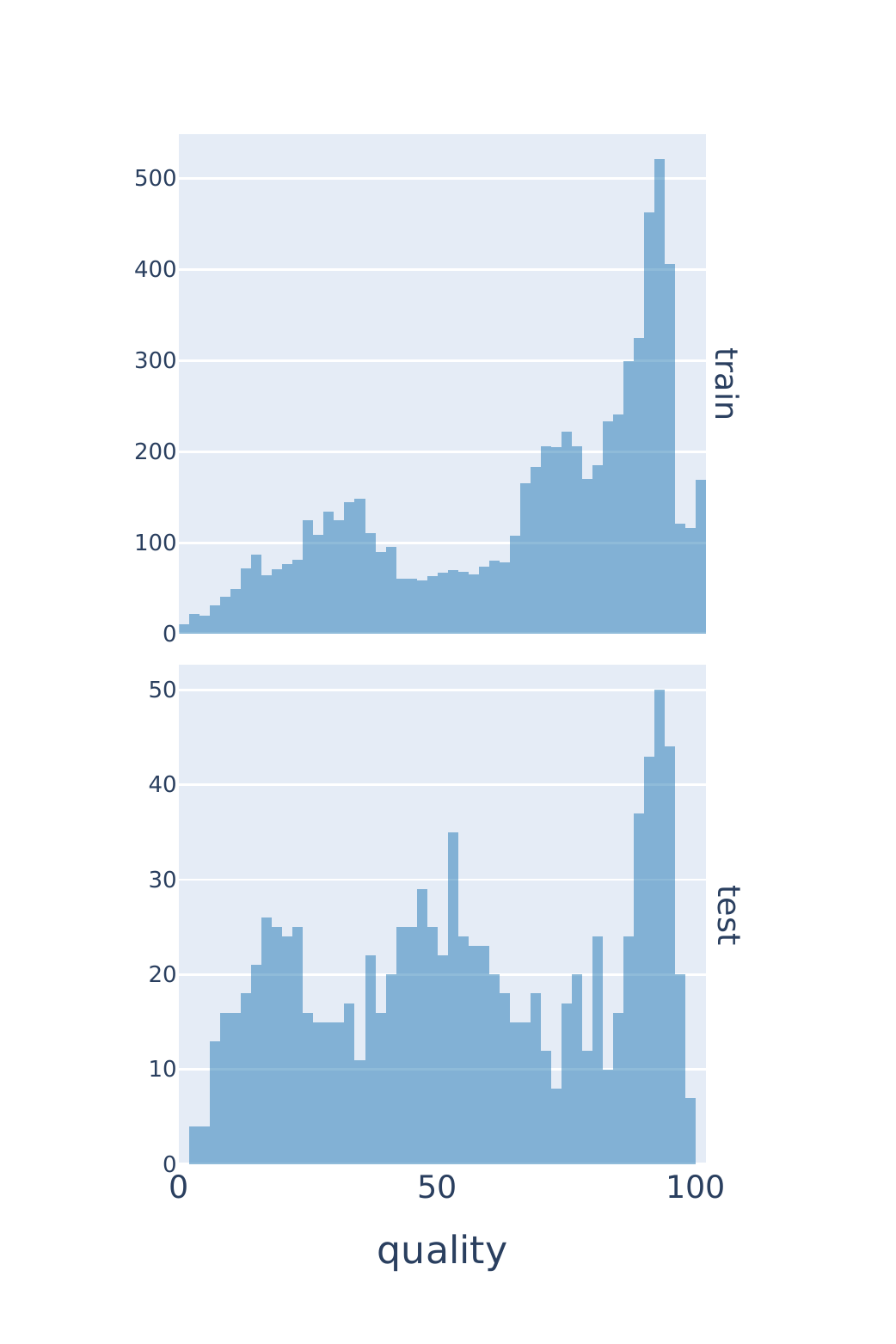}
            }
        \qquad 
        \subfigure{%
            \label{fig:distrib-shuffled}
            \includegraphics[scale=.66]{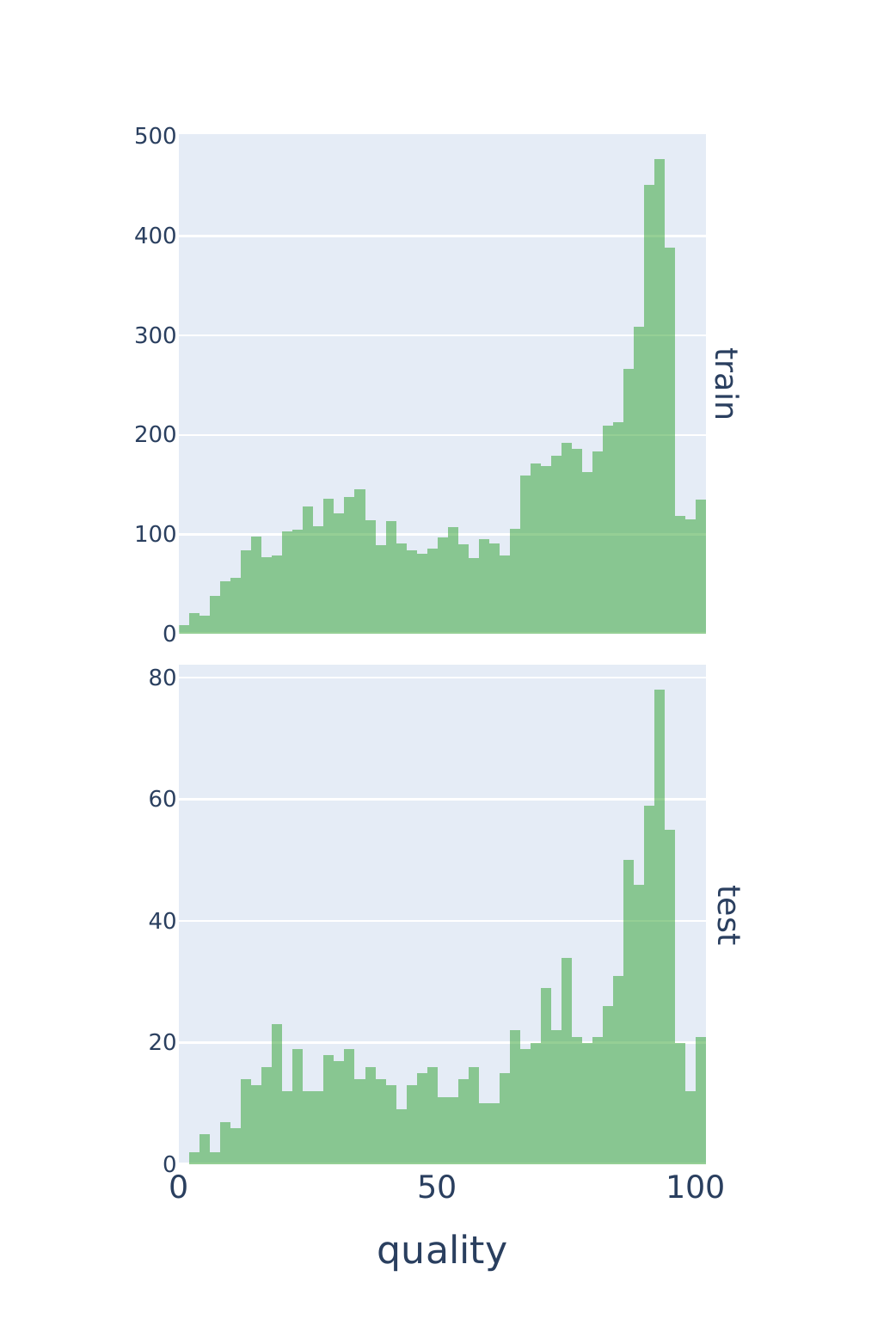}
        }
    }
\end{figure}

All datasets were randomly shuffled to ensure the independent and identically distributed (IID) assumption, which is crucial for conformal prediction as a framework in general. In Figure \ref{fig:distrib-original}, we plot the distribution of label values for training and test sets of the Estonian→English dataset, before shuffling. Figure \ref{fig:distrib-shuffled} shows the same quantities after randomly shuffling the original dataset. As can be seen, the shuffled version exhibits a higher similarity in label distribution between training and test set, indicating that the shuffling leads to a better IID property of the data.

Our experiments confirm the impact of shuffling on the performance of our models. The results show that our method consistently performs better on shuffled datasets, compared to datasets that were not shuffled. More details are included in Appendix \ref{apx:shuffling}.

We shuffle our datasets three times, each with a different random seed, and report the average performance of the models over the three versions of the dataset.

Our fine-tuning process is a three-epochs training of the pre-trained XLM-RoBERTa model over each dataset; we keep the model which scored the best Pearson's correlation on the validation set among the three epochs.

\subsection{Results and analysis}
In this section we demonstrate and analyse our results. Table \ref{tab:perf} summarises the performance of our models against the metrics described in section \ref{eva}. 

\begin{table*}
\centering
\begin{tabular}{llrrr}
\toprule
& & \textbf{\%ECE} & \textbf{Sha@90\%} & \textbf{AUC@10\%} \\
\midrule
\multirow{2}{5em}{En-De}    & \footnotesize\textsf{baseline}   & 13.88   & 2.81   & 0.63   \\
                            & \footnotesize\textsf{CPD}        & 2.06   & 2.29   & 0.62    \\
\midrule
\multirow{2}{5em}{En-Zh}    & \footnotesize\textsf{baseline}    & 3.19  & 2.51  & 0.78  \\
                            & \footnotesize\textsf{CPD}         & 1.06  & 2.48  & 0.78  \\
\midrule
\multirow{2}{5em}{Ro-En}  & \footnotesize\textsf{baseline} & 3.96  & 1.92 & 0.92  \\
                            & \footnotesize\textsf{CPD}    & 1.88   & 1.77  & 0.92 \\
\midrule
\multirow{2}{5em}{Et-En}  & \footnotesize\textsf{baseline} & 4.20   & 2.45  & 0.83 \\
                            & \footnotesize\textsf{CPD}    & 1.69   & 2.33  & 0.83  \\
\midrule
\multirow{2}{5em}{Si-En}  & \footnotesize\textsf{baseline} & 2.82   & 2.61  & 0.80 \\
                            & \footnotesize\textsf{CPD}    & 1.34   & 2.53  & 0.81   \\
\midrule
\multirow{2}{5em}{Ne-En}  & \footnotesize\textsf{baseline} & 3.11   & 2.40  & 0.79  \\
                            & \footnotesize\textsf{CPD}    & 1.62   & 2.39  & 0.80  \\
  
\bottomrule
\end{tabular}
\caption{
Performance averaged over three runs with different random train/validation/test splits. Our CPD-based model consistently outperforms the baseline in terms of expected calibration error and sharpness.
}
\label{tab:perf}
\end{table*}

  

\paragraph{Coverage} 
ECE measures how often the prediction intervals fail to include the true label, averaged over several significance levels $\epsilon\in\mathcal{E}$. For instance, an ECE of 1\% corresponds to a 1\% rate of the true label missing in the prediction interval, on average over $|\mathcal{E}|$ significance levels.

We note that the baseline model performs relatively well, apart from the English→German case. However, our model based on CPD consistently improves over the baseline by a factor of at least $2\times$ and does not suffer as much on the English→German dataset. The calibration error is always less or equal to 2\%, a good coverage result that may be improved upon with the use of more complex underlying algorithms or additional textual features.

It is important to highlight that these results are conditional to the datasets being randomly shuffled; in Appendix \ref{apx:shuffling} we show how CPD's performance is poorer whenever the IID assumption is less likely to hold. 

\begin{figure}[htbp]
\centering
\includegraphics[scale=0.6]{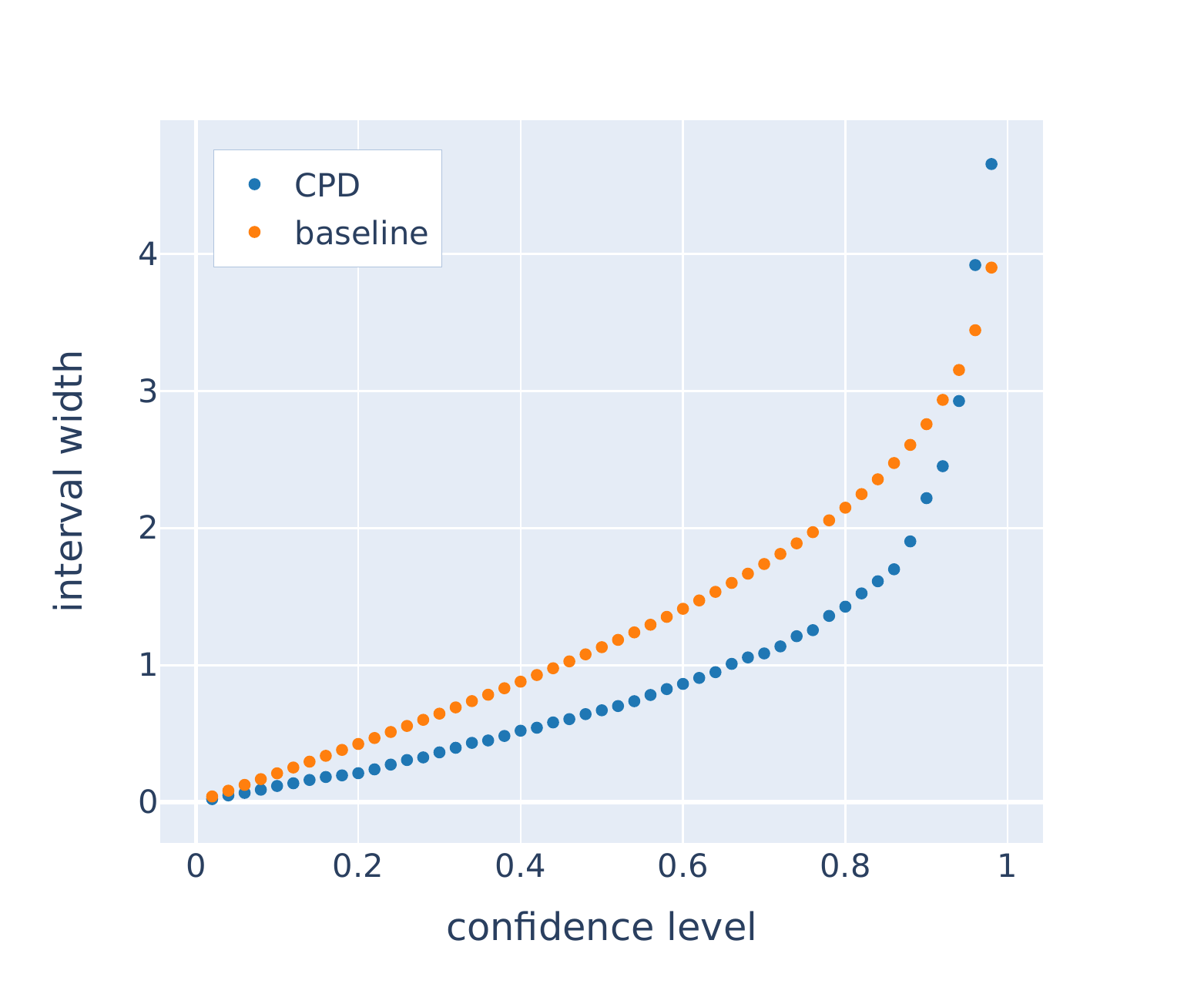}
\caption{Sharpness performance on the English→German dataset. Each point is the prediction interval size averaged over all test examples for a particular confidence level $1-\epsilon$. CPD intervals are tighter on average, then increase in size rapidly when approaching higher confidence levels.}
\label{fig:sharpness}
\end{figure}

\paragraph{Sharpness}
Again we verify a consistent improvement over the baseline. Differences are not as marked as they are in the coverage case, but in general we can say that at 90\% confidence ($\epsilon=0.1$), CPD produces tighter prediction intervals with better coverage. In Figure \ref{fig:sharpness} we have a detailed view of how prediction interval size varies with significance level, for both models. On the English→German dataset, we note that CPD widens its intervals rapidly when approaching high-confidence levels ($\epsilon<0.1$). At the upper limit of the significance range, baseline models exhibit greater sharpness than CPD. However, narrow intervals can be misleading if not paired to high-coverage predictions. We found this behaviour to be less prominent in the case of the other language pairs.

Note that the sharpness values in Table \ref{tab:perf} relate to the prediction intervals for the z-normalised quality scores $y$. For the Estonian→English dataset, for example, $q^*\in[0,100]$ corresponds to $y\in[-2.8, 1.4]$, and a prediction interval of width 2.46 would span a large quality range ($\hat q \in[20,90]$).

\paragraph{Detection of critical failures}
AUC@10\% is the AUROC score achieved by each model for the task of predicting if a translation is ``not good enough'' (specifically, if its quality score falls in the bottom decile). For this binary classification task, both baseline and CPD models achieve essentially the same results, both suffering on the English→German dataset and doing very well on Romanian→English.

\section{Conclusion and Future Work}
We presented a novel approach to quality estimation for MT based on conformal predictive distributions. Rather than returning a single quality score for each sentence pair, our model generates a prediction interval which is larger the more is the uncertainty of the prediction. More importantly, the predictions have guaranteed coverage under the IID assumption.

The prediction intervals obtained allow for a range of useful ``downstream'' tasks, such as deciding whether to publish a specific translation or not, or to inform users about the confidence in the quality estimate of a certain translation, or to rank translations produced by different MT models.

Our experimental results confirm the importance of the IID assumption for the successful application of conformal methods in NLP tasks. One potential avenue for future research is to explore methods for determining whether the IID assumption has been violated during the training process and at what point this occurred. \citet{pmlr-v152-vovk21b-retrain}'s recent research in this area is a promising step towards addressing this challenge.

\section*{Acknowledgements}
Thanks to Alex Gammerman and Ilia Nouretdinov for their constant support. Thanks to Chris Watkins and Alexander Balinsky for some insightful conversations. This work was partly supported by Centrica PLC.

\appendix

\section{Importance of shuffling}\label{apx:shuffling}
Table \ref{tab:perf-original} reports the performance over the original dataset splits. We can see that CPD does not always improve calibration over the Gaussian baseline. CPD does much better when we shuffle the datasets into new train/validation/test splits (Table \ref{tab:perf}).
\begin{table*}
\centering
\begin{tabular}{llrrr}
\toprule
& & \textbf{ECE} & \textbf{Sha@90\%} & \textbf{AUC@10\%} \\
\midrule
\multirow{2}{5em}{En-De}   & \footnotesize\textsf{baseline} & 6.56   & 2.24   & 0.637   \\
                            & \footnotesize\textsf{CPD}    & 3.75   & 1.99   & 0.642    \\
\midrule
\multirow{2}{5em}{En-Zh}    & \footnotesize\textsf{baseline}    & 1.31  & 2.21  & 0.728  \\
                            & \footnotesize\textsf{CPD}         & 1.58  & 2.17  & 0.727  \\
\midrule
\multirow{2}{5em}{Ro-En}  & \footnotesize\textsf{baseline} & 7.48  & 1.66 & 0.963  \\
                            & \footnotesize\textsf{CPD}    & 4.04   & 1.54  & 0.963 \\
\midrule
\multirow{2}{5em}{Et-En}  & \footnotesize\textsf{baseline} & 1.65   & 2.10  & 0.877 \\
                            & \footnotesize\textsf{CPD}    & 2.78   & 2.02  & 0.877  \\
\midrule
\multirow{2}{5em}{Si-En}  & \footnotesize\textsf{baseline} & 3.20   & 2.32  & 0.845 \\
                            & \footnotesize\textsf{CPD}    & 4.03   & 2.31  & 0.850   \\
\midrule
\multirow{2}{5em}{Ne-En}  & \footnotesize\textsf{baseline} & 5.23   & 2.04  & 0.883  \\
                            & \footnotesize\textsf{CPD}    & 3.29   & 1.97  & 0.871  \\
  
\bottomrule
\end{tabular}
\caption{
Performance over the original train/validation/test splitting.
}
\label{tab:perf-original}
\end{table*}

\bibliography{main}

\end{document}